\documentclass[letterpaper, 10 pt, conference]{ieeeconf}  % Comment this line out if you need a4paper
\usepackage[T1]{fontenc}
\usepackage{algorithm}% http://ctan.org/pkg/algorithms
\usepackage{xcolor}

\newif\ifshowdiffs
\showdiffsfalse        % <<<<<< FLIP THIS LINE
\ifshowdiffs
  \colorlet{diffcolor}{blue}
\else
  \colorlet{diffcolor}{black}
\fi
\DeclareRobustCommand{\diff}[1]{\textcolor{diffcolor}{#1}}
\usepackage{algpseudocode}% http://ctan.org/pkg/algorithmicx
\usepackage{ upgreek }
\usepackage{multicol}
\usepackage{ dsfont,setspace }
\usepackage{amsmath,graphicx}
\usepackage{amssymb}
\usepackage{amsmath}

\usepackage{lineno}
\usepackage{epstopdf}
\usepackage{authblk}
\usepackage{dblfloatfix}
\usepackage{tabularx, makecell}
\usepackage{booktabs}
\newcolumntype{L}{>{\raggedright\arraybackslash}X}
\usepackage[
  separate-uncertainty = true,
  multi-part-units = repeat
]{siunitx}
\let\labelindent\relax % ieeeconf.cls already defines \labelindent; avoids enumitem error
\usepackage{enumitem}
\usepackage[left=1.91cm, right = 1.91cm, top=2.02cm, bottom=1.91cm]{geometry}
\usepackage{subcaption}
\usepackage[colorlinks]{hyperref}
\hypersetup{
     colorlinks,
     linkcolor=black,
     citecolor=black,
     filecolor=black,
     urlcolor=blue,
 }
\usepackage{cleveref}
\usepackage{textcomp}
\usepackage{gensymb}
\usepackage{graphicx}

\IEEEoverridecommandlockouts                              % This command is only needed if 
\renewcommand{\arraystretch}{0.95}

\begin{document}
\bstctlcite{showAuthors}

\title{\LARGE \bf
\diff{BiPathCPG}: A Novel \diff{Load-Sharing} Design Architecture for Compound Planetary Gearboxes
}

\title{\LARGE \bf
\diff{BiPathCPG}: A Novel \diff{Load-Sharing} Design Architecture for Compound Planetary Gearboxes
}

% \author{
% % \thanks{This work is support by the DRDO Project.}
% %\thanks{$^{1}$A. Singh are with the Robert Bosch Center for Cyber Physical Systems, Indian Institute of Science, Bengaluru.}
% %\thanks{$^{2}$S. Kolathaya is with the Robert Bosch Center for Cyber Physical Systems and the Department of Computer Science \& Automation, Indian Institute of Science, Bengaluru.}

% }

\author{Deepak Kapa$^{*,1,2}$, Aman Singh$^{*,1}$, Saketh Vegesna$^{1}$,
Suryank Joshi$^{1}$,\\ Mayank Goswami$^{2}$ and Shishir Kolathaya$^{1}$%
% \thanks{*Authors contributed equally to this work.}%
% \thanks{This research is supported by ARTPARK at IISc.}%
% \thanks{$^{1}$Department of Cyber-Physical Systems (CPS), Indian Institute of Science (IISc), Bengaluru, India.
% {\tt\scriptsize \{saman, suryankjoshi, shishirk\}@iisc.ac.in}}%
% \thanks{$^{2}$Department of Physics, Indian Institute of Technology Roorkee, Roorkee, India.
% {\tt\scriptsize \{deepak\_k1, mayank.goswami\}@ph.iitr.ac.in}}%
% }
% \thanks{*Equal contribution. This work was supported by the DIA-RCoE under grant DFTM/02/3125/M/01/AIR-01. We thank Divyansh Rahangdale for his help with the experiments.}%
\thanks{*Equal contribution.}
\thanks{This work was supported by the DRDO Industry Academia Raman Centre of Excellence, IISc, under grant DFTM/02/3125/M/01/AIR-01.}
\thanks{We thank Divyansh Rahangdale for his help with the experiments.}%
\thanks{$^{1}$Dept. of Cyber-Physical Systems, Indian Institute of Science, Bengaluru, India. {\ttfamily\scriptsize\{saman,sakethappal1,suryankjoshi,shishirk\}@iisc.ac.in}}%
\thanks{$^{2}$Dept. of Physics, Indian Institute of Technology Roorkee, India. {\ttfamily\scriptsize\{deepak\_k1,mayank.goswami\}@ph.iitr.ac.in}}}

\maketitle
\thispagestyle{empty}
\pagestyle{empty}

\begin{abstract}
Quasi-Direct Drive (QDD) actuators are widely used in legged robots due to their low
reflected inertia, high backdrivability, and high torque density, commonly employing
planetary or cycloidal gearboxes. Single-stage planetary gearboxes offer high efficiency
but limited torque density; multi-stage designs improve torque density at the cost of
efficiency; and cycloidal gearboxes provide high stiffness but suffer from torque ripple.
Compound Planetary Gearboxes (CPG) offer a compromise, but conventional designs result
in increased actuator width, mass, and reduced torsional stiffness in 3D-printed
actuators. This paper proposes the \diff{BiPath Compound Planetary Gearbox
(BiPathCPG)}, a novel architecture that introduces \diff{an idle}
second-stage sun gear and load-sharing second-stage planets within the otherwise unused space of a
conventional CPG to simultaneously improve stiffness, reduce actuator width, and reduce
mass. An optimization framework computes \diff{BiPathCPG} parameters for a given
motor and target gear ratio. The architecture is validated through a 3D-printed
proof-of-concept actuator design, benchmarked against a conventional CPG using custom-built test rigs.
The \diff{BiPathCPG} achieves a torsional stiffness of \diff{1394}\,N$\cdot$m/rad, a \diff{$1.70\times$}
improvement over the conventional CPG, with \diff{a $32\%$} reduction in backlash, \diff{25}\,g
lower mass, and approximately \diff{$6.4$}\,mm reduction in axial width, while \diff{achieving
a peak stall torque of $36.49$\,N$\cdot$m}. \href{https://deepak-kapa.github.io/PancakeActuator/}{[Webpage]}
\end{abstract}

%%%%%%%%%%%%%%%%%%%%%%%%%%%%%%%%%%%%%%%%%%%%%%%%%%%%%%%%%%%%%%%%%%%%%%%%%%%%%%%%

\section{Introduction}

Industrial demand for robots, including humanoids and quadrupeds, is increasing. These robots require efficient, high-performance joint actuators, often with metal gearboxes~\diff{\cite{garcia_cyc_bad}}, \cite{10.3389/frobt.2018.00129}.
% Several actuator types exist for this purpose~\cite{10.3389/frobt.2020.00103, 10.3389/frobt.2018.00129}.
At the same time, growing interest in cost-effective legged robot research platforms~\cite{chi2025demonstratingberkeleyhumanoidlite, solo} has increased interest in actuators with 3D-printed gearboxes~\cite{anti_backlash_cyc, embiract,bezzini2024gearboxes}. In both scenarios, one of the most widely adopted solutions is the Quasi-Direct Drive (QDD) actuator~\cite{mitqqd1, minicheetah, Zhu2023Humanoid, sspg_qdd, IeCAT2020-08516}, which pairs a high torque-density motor with a low gear ratio, enabling low reflected inertia and high backdrivability.

\begin{figure}[htbp]
    \centering
    \includegraphics[width=0.75\linewidth]{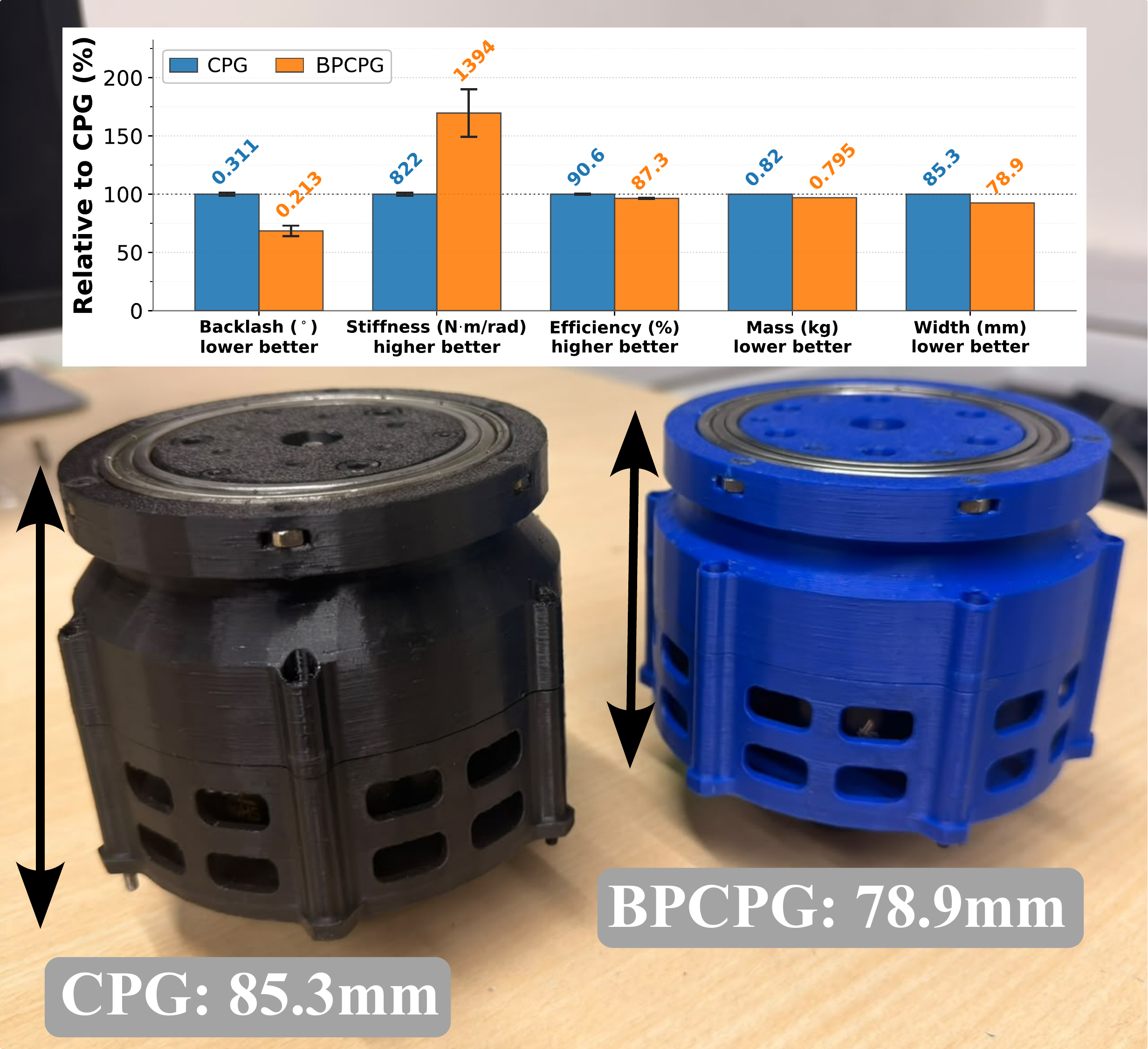}
    \caption{\textbf{Actuators (3D-printed)}: Compound Planetary Gearbox (CPG) (left, \diff{820}\,g) and \diff{BiPathCPG} (right, \diff{795}\,g).}
    \label{fig:intro}
\end{figure}

In metal-gearbox QDD actuators, one of the most commonly used transmission architectures is the single-stage planetary gearbox (SSPG)~\cite{mitcheetah3design, sspg_qdd, mitqqd1, minicheetah}. SSPGs are efficient and compact, but their torque density is limited by achievable gear ratios and peak torque capacity. Alternative planetary architectures, such as the Wolfrom planetary gearbox (WPG, 3K)~\cite{choe2025design3dofhoppingrobot}, \diff{\cite{bezzini2026embedded}} and double-stage planetary gearbox (DSPG)~\cite{dspg_qdd}, provide higher torque density through larger gear ratios, but WPGs typically exhibit lower efficiency and DSPGs suffer reduced efficiency from multi-stage power losses.
% Cycloidal gearboxes have also been explored for QDD actuators~\cite{cqdd}, but they are known to exhibit significant output torque ripple~\cite{output_wiko_2019}.

% Furthermore, both WPG and DSPG configurations generally result in heavier gearboxes.
% If we consider QDD actuators with 3D-printed gearboxes, planetary gearboxes are often at a disadvantage compared to cycloidal gearboxes~\cite{cqdd, 3dp_planetary_breakage, embiract}. The cycloidal gearboxes are much more axially compact as compared to planetary gearboxes~\cite{pinwheel, roozing2022cycloidal, bezzini2024gearboxes}. There are several other works as well that build cycloidal-planetary hybrid gearboxes, that are also very axially compact~\cite{bezzini2025cyplo, bezzini2026nested}. The primary reason is that the cycloidal gearboxes provide high gear-ratios compared to planetary gearboxes in a given space. This is primarily due to the lower strength of 3D-printed materials, which requires wider gears and larger gearbox dimensions to meet strength requirements, resulting in increased actuator width and mass. Additionally, the stiffness of 3D-printed planetary actuators is typically lower than that of 3D-printed cycloidal actuators~\cite{chi2025demonstratingberkeleyhumanoidlite, singh2025compactcomputationaloptimizationautomated, anti_backlash_cyc}. However, cycloidal gearboxes suffer from output torque ripple~\cite{output_wiko_2019, garcia_cyc_bad}.

For QDD actuators with 3D-printed gearboxes, planetary gearboxes are often at a disadvantage compared with cycloidal gearboxes~\cite{cqdd, 3dp_planetary_breakage, embiract}, \diff{which are typically more axially compact~\cite{pinwheel, roozing2022cycloidal, bezzini2024gearboxes}. Axially compact cycloidal--planetary hybrid gearboxes have also been developed~\cite{bezzini2025cyplo, bezzini2026nested}. Cycloidal gearboxes achieve higher gear ratios within a given space, while} the lower strength of 3D-printed materials requires wider planetary gears and larger dimensions, increasing actuator width and mass. 3D-printed planetary actuators also exhibit lower stiffness than 3D-printed cycloidal actuators~\cite{chi2025demonstratingberkeleyhumanoidlite, singh2025compactcomputationaloptimizationautomated, anti_backlash_cyc}. Cycloidal gearboxes, however, can exhibit output torque ripple~\cite{output_wiko_2019}, \diff{\cite{garcia_cyc_bad}}. \diff{While we recognize the advantages of cycloidal architectures in axial compactness, this work focuses on addressing the limitations of 3D-printed planetary gearboxes.}

Among planetary architectures, compound planetary gearboxes (CPGs) offer an effective compromise between efficient but low-torque-density SSPGs and high-torque-density but less efficient WPGs and DSPGs~\cite{8202172, liao2024berkeleyhumanoidresearchplatform}. CPGs (Fig.~\ref{fig:designs}(a)--(c)) \diff{achieve higher gear ratios than SSPGs} at comparable efficiency~\cite{singh2025compactcomputationaloptimizationautomated}. However, the smaller second-stage planet diameters require relatively wide second-stage planet and ring gears, increasing actuator mass in both metal and 3D-printed CPGs. 3D-printed CPG actuators also exhibit lower stiffness than 3D-printed cycloidal gearbox actuators~\cite{chi2025demonstratingberkeleyhumanoidlite, singh2025compactcomputationaloptimizationautomated, anti_backlash_cyc}.

To address the large width and mass of compound planetary actuators (CPG), as well as the lower stiffness observed in 3D-printed CPG actuators, we propose a new design called the \diff{\textbf{BiPath Compound Planetary Gearbox (BiPathCPG)}}. This proof-of-concept architecture for compound planetary gearboxes employs \diff{idle} load-sharing gears positioned within the unused space of a traditional CPG configuration, achieving stiffness and backlash characteristics comparable to cycloidal gearboxes while maintaining high efficiency.
% low wear over time, and low gear inertia that are typical advantages of planetary actuators~\cite{Katz2018Actuator}.
Our main contributions are:
\begin{itemize}
\item We propose a novel compound planetary gearbox architecture, the \diff{BiPathCPG}, which improves stiffness and reduces actuator width and mass.

\item We experimentally validate the design through a 3D-printed proof-of-concept prototype and compare it with a traditional compound planetary gearbox.

\item We develop a Python-based framework that computes the \diff{BiPathCPG} design variables for a given motor and desired gear ratio. \href{https://anonymous.4open.science/r/PancakeActuator-2802/README.md}{[Code-link]}

\end{itemize}

\section{Background}

\subsection{Compound Planetary Gearbox (CPG) Architecture}

\begin{figure}[htbp]
    \centering
    \includegraphics[width=\columnwidth]{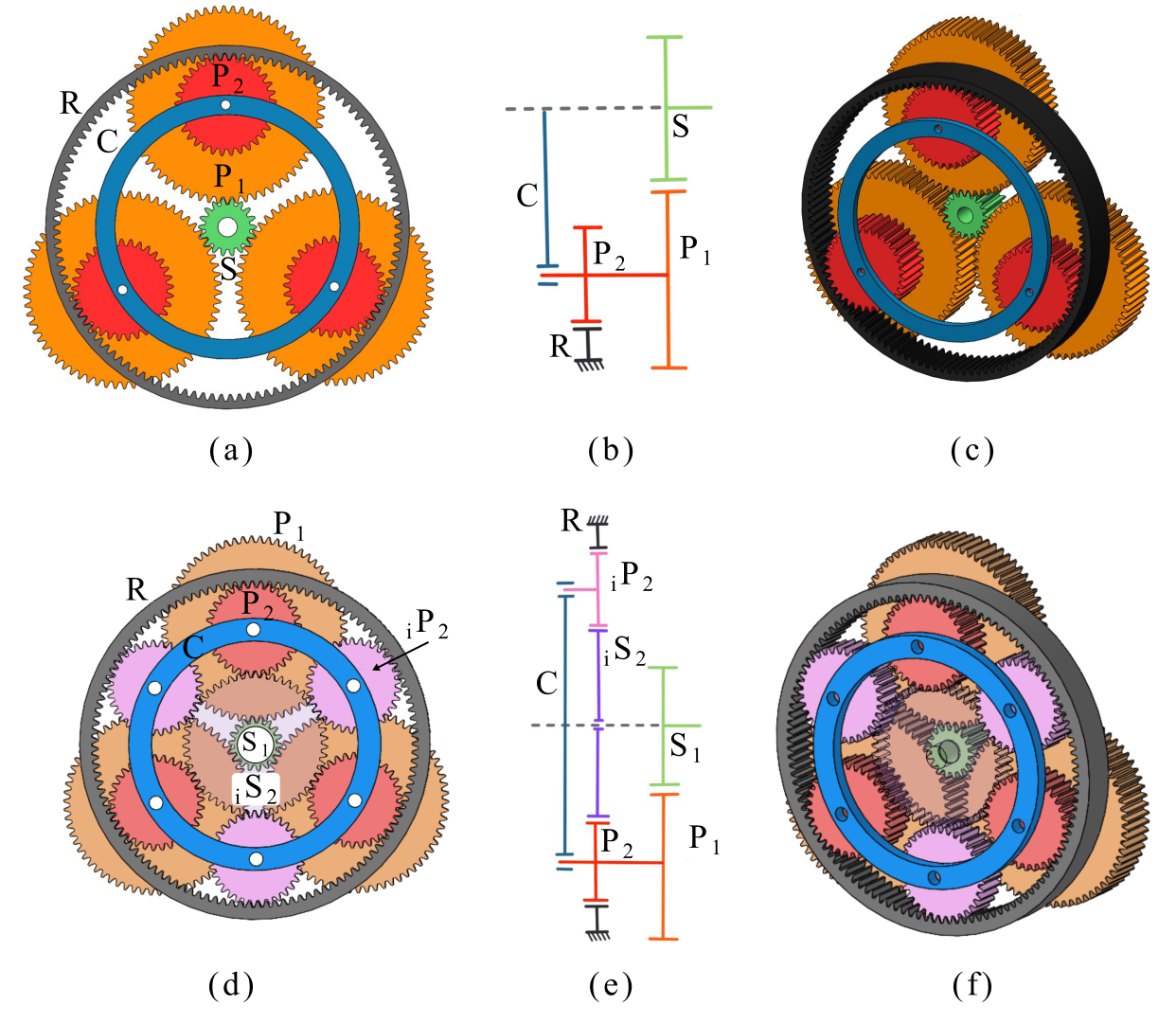}
    \caption{\textbf{Gearbox architectures}: (a)--(c) CPG and \diff{(d)--(f) BiPathCPG},
    shown as front view, line diagram and isometric view, respectively.
    \diff{$S$: sun, $P_1$/$P_2$: first-/second-stage planets, $R$: ring, $C$: carrier.}
    \diff{$iS_2$ and $iP_2$ denote the idle second-stage sun and planet gears of the BiPathCPG.}}
    \label{fig:designs}
\end{figure}

The compound planetary gearbox (CPG) is a planetary architecture that uses compound
planets: two gears of different diameters rigidly connected to rotate at the same angular
velocity. In the most torque-dense configuration, the sun drives the larger planet, the
smaller planet meshes with a stationary ring, and the output is taken from the carrier.
Fig.~\ref{fig:designs}(a)--(c) shows a 2D front view and line diagram of this configuration.

The gear reduction ratio is given by

\begin{equation}
G_{\mathrm{CPG}} 
=
\frac{\left(N_{s1} + N_{p1}\right)
      \left(N_{p2} + \frac{m_1}{m_2} N_{p1}\right)}
     {\left(N_{s1}\right)\left(N_{p2}\right)},
\label{eq:CPG_ratio}
\end{equation}

where $N_{s1}$ is the number of teeth on the sun gear, $N_{p1}$ and $N_{p2}$ those on the
larger and smaller planet gears, and $m_1$ and $m_2$ the modules of the sun--planet and
planet--ring meshes\diff{, taken equal ($m_1 = m_2$) for simplicity}.
\diff{The overall efficiency is
\begin{equation}\label{eq:cpg_eff}
\eta_{\text{CPG}} =
\frac{
N_{s_1}N_{p_2}
+ \eta_{s_1p_1}\eta_{p_2r_2}N_{p_1}N_{r_2}
}
{(N_{s_1}+N_{p_1})(N_{p_2}+N_{p_1})},
\end{equation}
where $\eta_{s_1p_1}$ and $\eta_{p_2r_2}$ are the sun--planet and planet--ring mesh
efficiencies, calculated using the formulation in \cite{effModel3K}.}

% Compared to a single-stage planetary gearbox, the CPG achieves substantially higher reduction ratios within a similar radial footprint while maintaining comparable efficiency (with a marginal reduction due to additional meshing losses), thereby enabling high torque density suitable for legged locomotion applications.

% \textcolor{red}{A key qualitative design consideration arises from torque transmission through the compound planet assembly. The second (smaller) planet gear, which forms the second reduction stage, is subjected to high contact forces. However, due to its smaller pitch diameter, the transmitted torque results in elevated tangential tooth forces. To satisfy bending strength and contact stress constraints, an increased face width is typically required. This directly increases the axial length of the actuator, potentially leading to reduced volumetric efficiency.}

% \begin{figure}[htbp]
%     \centering
%     \includegraphics[width=0.9\linewidth]{images/Lewis_equation_diagram.png}
%     \caption{}
%     \label{fig:lewis_eq_diag}
% \end{figure}

% \begin{figure*}[htbp]
%     \centering
%     \includegraphics[width=0.8\textwidth]{images/cpg_pancake_Lewis_equation_diagram.png}
%     \caption{\textcolor{red}{Flip the arrow sign of $\tau_{out}$, make it $\tau_{load}$, add a dotted circle to the carrier and show it $\tau_{load}$ on that circle}}
%     \label{fig:cpg_lewis_eq_diag}
% \end{figure*}

\subsection{Gear Tooth Strength Modeling}

Gear width is an important gearbox design parameter. It depends on the operating load. We determine it using the modified Lewis gear strength equation~\cite{maitra1994gear}, as follows:

\begin{equation}
     \sigma_b = \frac{F_t}{b \cdot m} q_e q_k
     \label{eq:lewis_eq}
\end{equation}  

To calculate the face-width of the gear we can rearrange the Eq.~\ref{eq:lewis_eq} and add a Factor of Safety ($FOS$), as follows:

\begin{equation}
     b = FOS \cdot \frac{F_t}{\sigma_b \cdot m} q_e q_k
     \label{eq:face_width_exp}
\end{equation} 

where $\sigma_b$ is the bending strength of the gear material, $F_t$ is the tooth force applied to the gear tooth, $b$ is the width (face-width) of the gear, $m$ is the module of the gear, $q_e = \frac{1}{CR}$ is the factor from contact ratio, and $q_k$ is the tooth form factor, calculated using the figure given in \cite{maitra1994gear}, \diff{\href{https://drivetrainhub.com/notebooks/gears/strength/Chapter\%202\%20-\%20Root\%20Stress.html}{[Drivetrain Hub]}}. Here, $FOS=2$.

% So, we can see that the face-width of the gear is dependent on the tooth force it is undergoing, the bending strength of the material, the module and the contact ratio form factor and the tooth form factor, which depends upon the number of teeth of the gear.

% \textcolor{red}{
\section{Problem Formulation}\label{sec:Problem_Formulation}
\subsection{Increased Mass and width of the CPG Actuator}\label{cpg_issues}
\subsubsection{Static Force Modeling of the Compound Planetary Gearbox (CPG)}
\begin{figure}[htbp]
    \includegraphics[width=\columnwidth]{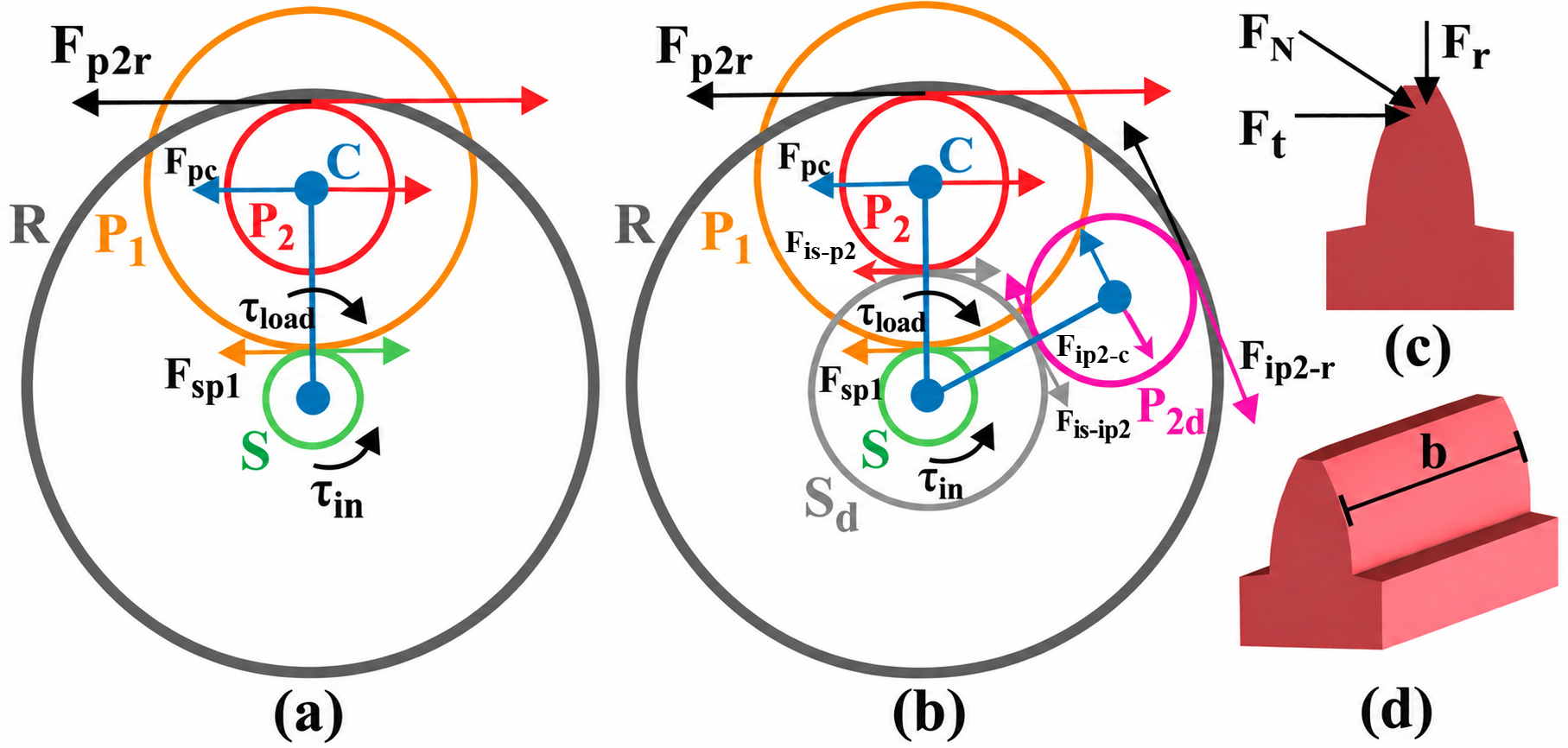}
    \caption{Free body diagrams of (a) CPG and (b) \diff{BiPathCPG}; (c)–(d) gear tooth \diff{tangential} contact forces.}
    \label{fig:cpg_lewis_eq_diag}
\end{figure}

Gear face width is set by the tooth forces, which follow from a steady-state
force balance: the motor applies $\tau_{in}$ to the sun gear while the load at
the carrier holds every gear at constant angular velocity. From the free-body
diagram in Fig.~\ref{fig:cpg_lewis_eq_diag}, the equilibrium equations are

\begin{equation}
\begin{bmatrix}
n_{p} & 0 & 0 \\
0 & n_{p} & 0 \\
r_{p1} & 0 & r_{p2}
\end{bmatrix}
\begin{bmatrix}
F_{sp1} \\
F_{pc} \\
F_{p2r}
\end{bmatrix}
=
\begin{bmatrix}
\frac{\tau_{in}}{r_s} \\
\frac{\tau_{load}}{(r_s + r_{p1})} \\
0
\end{bmatrix}
\end{equation}

where:
\begin{itemize}[leftmargin=*,noitemsep,topsep=2pt]
  \item $F_{sp1}$, $F_{p2r}$, $F_{pc}$: \diff{tangential} contact forces at the sun--planet~1, planet~2--ring, and planet--carrier interfaces
  \item $r_s, r_{p1}, r_{p2}$: radii of the sun, planet~1 (larger) and planet~2 (smaller) gears
  \item $n_p$: number of planet gears
  \item $\tau_{in}, \tau_{load}$: motor input torque and output load torque
\end{itemize}
Under steady-state operation, with all gears at constant angular velocity, the load torque satisfies
\diff{\begin{equation}
\tau_{load} = - G \, \tau_{in},
\end{equation}}
where $G$ denotes the gear ratio. 
\diff{
For simplicity, transmission losses are neglected; therefore, efficiency is not included in the output torque calculation.
}
To determine the maximum tangential forces acting on the gear teeth, the input torque $\tau_{in}$ is taken as the peak torque rating of the motor used in the actuator.

Solving these equations gives

\begin{equation}
F_{sp1}=\frac{\tau_{in}}{n_pr_s},\quad
F_{pc}=\frac{G\tau_{in}}{n_p(r_s+r_{p1})},\quad
F_{p2r}=-\frac{r_{p1}}{r_{p2}}\frac{\tau_{in}}{n_pr_s}.
\label{eq:force_sol}
\end{equation}

% From \eqref{eq:Fp2r}, it can be observed that the force between the second-stage planet gear (planet–2) and the ring gear, $F_{p2r}$, is $(r_{p1}/r_{p2})$ times the magnitude of $F_{sp1}$. Since $r_{p1} > r_{p2}$, it follows that $F_{p2r} > F_{sp1}$. Additionally, as reported in~\cite{maitra1994gear, eritenel2019gearrootstress}, the factor $q_k$ increases with a decreasing number of teeth. According to \eqref{eq:lewis_eq}, these two factors together result in a significantly larger required face width for the second-stage planet gear. Since the ring gear must have the same face width as the meshing planet gear, the ring gear width also increases. This increase in gear width leads to an increase in the overall actuator width and mass, as other components such as the carrier and outer housing must also be sized accordingly.

From \eqref{eq:force_sol}, $F_{p2r}$ is $(r_{p1}/r_{p2})$ times the magnitude of $F_{sp1}$. Since $r_{p1}>r_{p2}$, $F_{p2r}>F_{sp1}$. Moreover, $q_k$ increases as the number of teeth decreases~\cite{maitra1994gear}. According to \eqref{eq:lewis_eq}, these factors require a significantly larger face width for the second-stage planet gear. Since the ring gear must match the face width of the meshing planet gear, its width also increases. This, in turn, increases the overall actuator width and mass, as components such as the carrier and outer housing must be sized accordingly.

\subsection{\diff{Stiffness Modeling of CPG Gearbox}}\label{sec:stiffness_problem}

\diff{
To estimate the mesh-induced stiffness of the CPG, we adopt a simplified
lumped-parameter formulation similar to~\cite{stiffnessModel2}, in which
each gear mesh is represented by a linear spring acting along the tangential
direction. Only the compliance of the gear meshes is considered, while the
deformation of the gears, planet shafts, carrier, bearings, and housing is neglected.
The total elastic strain energy stored in the CPG meshes is expressed as
\begin{equation}
    U_{\mathrm{CPG}}
    =
    \sum_{i \in \{sp1,p2r\}}
    \frac{F_i^2}{2k_i},
\end{equation}
where $k_{sp1}$ and $k_{p2r}$ denote the linear mesh stiffnesses of the
sun--planet-1 and planet-2--ring meshes, respectively
~\cite{strainEnergyUsage}.
Using the mesh forces obtained in~\eqref{eq:force_sol}, the input angular
deflection is obtained from Castigliano's theorem~\cite{hibbeler_mechanics}
as
\begin{equation}
    \theta_{\mathrm{in}}
    =
    \frac{\partial U_{\mathrm{CPG}}}
         {\partial \tau_{\mathrm{in}}}.
\end{equation}
The resulting input-referred torsional stiffness is
\begin{equation}
    K_{\mathrm{CPG,in}}
    =
    \frac{\tau_{\mathrm{in}}}{\theta_{\mathrm{in}}}
    =
    n_p r_s^2
    \left[
        \frac{1}{k_{sp1}}
        +
        \left(
            \frac{r_{p1}}{r_{p2}}
        \right)^2
        \frac{1}{k_{p2r}}
    \right]^{-1}.
\end{equation}
The corresponding output-referred stiffness is
\begin{equation}
    K_{\mathrm{CPG,out}}
    =
    G^2 K_{\mathrm{CPG,in}}.
\end{equation}
}
\diff{
Since $r_{p1}>r_{p2}$ in a CPG, the contribution of the second-stage mesh
compliance is amplified by the factor $(r_{p1}/r_{p2})^2$. Therefore, for
comparable mesh stiffnesses, the CPG exhibits lower torsional stiffness than
a single-stage planetary gearbox, for which $r_{p1}=r_{p2}$.
}

% Furthermore, the stiffness of a 3D-printed compound planetary gearbox is lower than that of a 3D-printed single-stage planetary gearbox (SSPG)~\cite{singh2025compactcomputationaloptimizationautomated}. The primary reason for the reduced stiffness in CPG gearboxes is that, under high torque conditions, the second-stage planet gear experiences both tangential and radial forces from the ring gear, as illustrated in Fig.~\ref{fig:cpg_lewis_eq_diag}(c). The tangential force causes sideways tooth deflection, while the radial force induces inward bending of the second-stage planet gear toward the sun gear. This deformation reduces the effective line contact between meshing gear teeth and may lead to partial loss of contact, particularly near the tooth-tip region. Additionally, due to this inward deflection, the tangential forces are transmitted farther from the tooth root, increasing the bending moment on the gear teeth. The increased bending moment leads to larger tooth deflection, which consequently reduces the overall transmission stiffness.

\section{BiPathCPG Gearbox}

To address the issues discussed in the previous section, we introduce the
\diff{BiPathCPG}. \diff{Idle} planet
gears are added to the second stage, Fig.~\ref{fig:designs}(d)--(f) (pink),
equal in number to the planets of a conventional CPG and identical to the
second-stage planets in diameter and tooth count. An \diff{idle}
second-stage sun gear is also added, Fig.~\ref{fig:designs}(d) (transparent);
it is not rigidly attached to the first-stage sun but rotates about the gearbox centre on a bearing and meshes with all second-stage planets, redundant ones included.
The idle planets open a second load path in the second stage, reducing the
force carried by each primary ring--planet mesh and so allowing a smaller
stage-2 face width, and with it a narrower and lighter actuator. 
% The redundant sun gear is intended to constrain inward deflection of the planet gears under high torque and thereby preserve line contact with the ring gear, an effect expected to matter most in 3D-printed gearboxes, whose gears are considerably more compliant than metal ones.

\subsection{Gear-Width Analysis of the \diff{BiPathCPG}}\label{force_assumption}

The static force equilibrium of the \diff{BiPathCPG} gearbox can be written as:

\begin{equation}
\label{eq:static_pancake}
\resizebox{\columnwidth}{!}{$
\begin{bmatrix}
n_{p} & 0 & 0 & 0 & 0 & 0 & 0 \\
0 & 0 & n_{p} & 0 & 0 & 0 & n_{p} \\
0 & 0 & 0 & n_{p} & n_{p} & 0 & 0 \\
r_{p1} & r_{p2} & 0 & r_{p2} & 0 & 0 & 0 \\
1 & -1 & -1 & 1 & 0 & 0 & 0 \\
0 & 0 & 0 & 0 & 1 & 1 & 0 \\
0 & 0 & 0 & 0 & 1 & -1 & -1 
\end{bmatrix}
\begin{bmatrix}
F_{sp1} \\
F_{p2r} \\
F_{pc} \\
F_{is-p2} \\
F_{is-ip2} \\
F_{ip2-r} \\
F_{ip2-c}
\end{bmatrix}
=
\begin{bmatrix}
\dfrac{\tau_{in}}{r_s} \\
\dfrac{\tau_{load}}{r_s + r_{p1}} \\
0 \\
0 \\
0 \\
0 \\
0
\end{bmatrix}
$}
\end{equation}

where:
\begin{itemize}[leftmargin=*,noitemsep,topsep=2pt]
  \item $F_{sp1}$, $F_{p2r}$, $F_{pc}$: contact forces at the sun--planet~1, planet~2--ring, and planet--carrier interfaces
  \item $F_{is\text{-}p2}$, $F_{is\text{-}ip2}$, $F_{ip2\text{-}r}$, $F_{ip2\text{-}c}$: contact forces involving the \diff{idle} sun and \diff{idle} planet gears (idle sun--planet~2, planet~2--idle planet, idle planet--ring, idle planet--carrier)
  \item $r_s, r_{p1}, r_{p2}, r_{is}, r_{ip2}, r_r$: radii of the sun, planet~1, planet~2, \diff{idle} sun, \diff{idle} planet, and ring gears
  \item $\tau_{in}, \tau_{load}$: motor input torque and output load torque
\end{itemize}

Equation~\eqref{eq:static_pancake} can be written compactly as
\[
A f = \tau,
\]
where $A \in \mathbb{R}^{7\times7}$, $f \in \mathbb{R}^{7\times1}$, and $\tau \in \mathbb{R}^{7\times1}$. The matrix $A$ is rank deficient, implying that the system admits infinitely many solutions. This occurs because the \diff{BiPathCPG} gearbox contains kinematically redundant gear interactions. 

% An analogous situation arises in statics when determining the reaction forces of a four-legged table supporting an asymmetric load. Without modelling compliance or contact properties, the equilibrium equations alone cannot uniquely determine the individual forces. Similarly, in the Pancake-CPG gearbox, additional physical assumptions are required to obtain a unique force distribution.

A similar case occurs in a four-legged table with an uneven load: equilibrium alone cannot uniquely determine each leg’s force without considering stiffness or contact model. Likewise, the \diff{BiPathCPG} gearbox needs additional physical assumptions to determine a unique force distribution.

\subsubsection{Load-sharing assumption}
To resolve this indeterminacy, we introduce the load-sharing relation
\begin{equation}
F_{ip2-r} = \gamma\,F_{p2r},
\label{eq:gamma}
\end{equation}
where $\gamma \in \mathbb{R}^{+}$ is the force-sharing factor between the
planet--ring and idle-planet--ring contacts. Although the idle planets are
kinematically redundant, $F_{ip2-r}$ is generally nonzero: \diff{gear-body and tooth}
compliance, bearing friction, manufacturing tolerances and profile deviations
distribute load over all stage-2 meshes. Thus $\gamma<1$ indicates that the idle
mesh carries less load than the primary planet mesh, $\gamma>1$ more, \diff{and the
limiting case $\gamma=0$ recovers the conventional CPG, in which the entire ring
reaction is carried by the planet meshes alone.}

Applying~\eqref{eq:gamma} reduces~\eqref{eq:static_pancake} to
\begin{equation}
\begin{bmatrix}
n_{p} & 0 & 0 & 0 & 0 & 0 \\
0 & 0 & n_{p} & 0 & 0 & n_{p} \\
0 & 0 & 0 & n_{p} & n_{p} & 0 \\
r_{p1} & r_{p2} & 0 & r_{p2} & 0 & 0 \\
1 & -1 & -1 & 1 & 0 & 0 \\
0 & \gamma & 0 & 0 & 1 & 0
\end{bmatrix}
\begin{bmatrix}
F_{sp1} \\ F_{p2r} \\ F_{pc} \\ F_{is-p2} \\ F_{is-ip2} \\ F_{ip2-c}
\end{bmatrix}
=
\begin{bmatrix}
\frac{\tau_{in}}{r_s} \\ \frac{\tau_{load}}{(r_s + r_{p1})} \\ 0 \\ 0 \\ 0 \\ 0
\end{bmatrix}
\label{eq:static_pancake_with_assumption}
\end{equation}
\diff{Solving it further, we get:
\begin{equation}
\begin{gathered}
F_{sp1}=\frac{\tau_{in}}{n_{p} r_s},\qquad
F_{p2r}=-\frac{1}{1+\gamma}\,\frac{r_{p1}}{r_{p2}}\,\frac{\tau_{in}}{n_{p} r_s},\\[2pt]
F_{is-p2}=-F_{is-ip2}=\gamma F_{p2r},\qquad
F_{ip2-r}=\gamma F_{p2r},\\[2pt]
F_{pc}=\left(1+\frac{1-\gamma}{1+\gamma}\frac{r_{p1}}{r_{p2}}\right)\frac{\tau_{in}}{n_{p} r_s},\\[2pt]
F_{ip2-c}=\frac{1}{n_{p}}\frac{\tau_{load}}{r_s+r_{p1}}-F_{pc}.
\end{gathered}
\label{eq:solution_gamma}
\end{equation}
The total ring reaction $F_{p2r}+F_{ip2-r}=(1+\gamma)F_{p2r}$ is independent of $\gamma$
and equals that of a conventional CPG: the idle planets just redistribute it over $2n_p$ meshes. Since the face width of each mesh is sized by its own
 contact load, this redistribution makes the second stage axially thinner.}
 % \textcolor{red}{Second, the same holds at the carrier: the total planet--carrier reaction
% $F_{pc}+F_{ip2-c}$ is fixed by $\tau_{load}$, while the share carried by the
% compound-planet pin, $F_{pc}$, decreases monotonically with $\gamma$. The
% output-path load is therefore split between the compound-planet and
% idle-planet pins in the same way, and again in a single axial plane.}}

\begin{figure}[htbp]
    \centering
    \includegraphics[width=\linewidth]{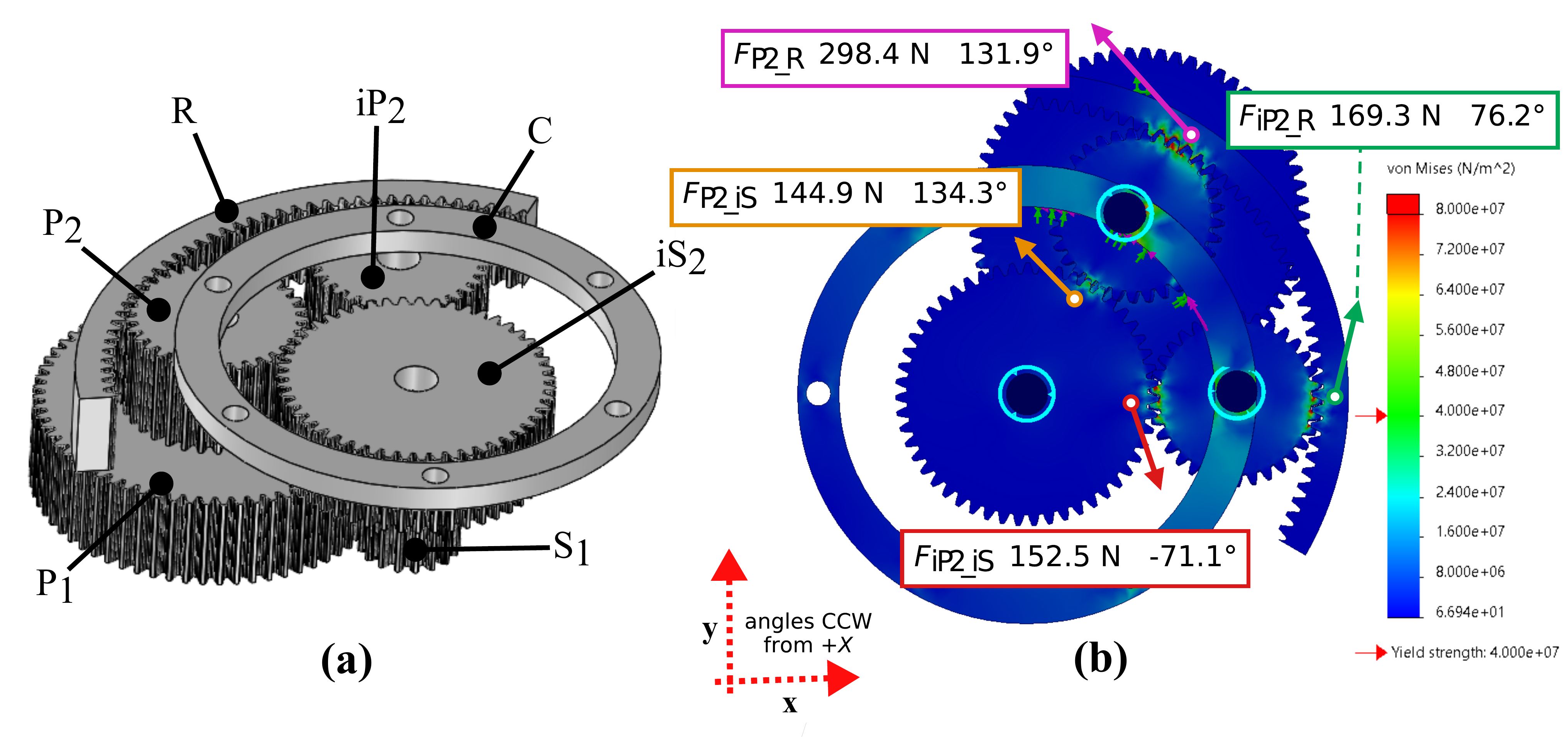}
    \caption{\diff{\textbf{FEA study of stage-2 load sharing.}}}
    % (a) Simplified gearbox model: stage-1 sun $S_1$, compound planet
    % $P_1$--$P_2$, idle planet $iP_2$, stage-2 idle sun $iS_2$, ring $R$ and
    % carrier $C$.
    % (b) von Mises stress and resultant ring-mesh contact forces for the
    % load-matched face widths, giving $\gamma = 189.8/332.6 = 0.571$.}
    \label{fig:fea}
\end{figure}

\subsubsection{\diff{FEA estimation of the force-sharing factor}}
\diff{A simplified model retaining one set of meshing gears
(Fig.~\ref{fig:fea}a) was analysed in SolidWorks Simulation: the compound
planets $P_1$--$P_2$, the idle planets $iP_2$ and the stage-2 idle sun $iS_2$
rotate freely, the stage-1 sun $S_1$ and the ring $R$ are fixed, $15$~N$\cdot$m
is applied to the carrier $C$, and all meshes start at zero backlash with
frictionless contact. With equal planet and idle-planet face widths of
$8.2$~mm the ring-mesh forces give $\gamma=0.566$; rescaling the face widths in
proportion to those loads (Fig.~\ref{fig:fea}b) gives $\gamma=0.5673$, and a
third study with the carrier fixed and $3$~N$\cdot$m applied to the sun
reproduces the same ratio. Because the model is frictionless this is a lower
bound: the idle path closes through the freely rotating $iS_2$, and bearing
friction there raises the torque carried along the idle branch. We therefore
adopt $\gamma = 0.6$.}

\diff{With $\gamma=0.6$ the planet--ring contact force is $62.5\%$ of
the value in a conventional CPG of identical gear geometry, the idle mesh
carrying the remaining $37.5\%$. The stage-2 planet face width, and hence the
axial width of the ring and of the outer shell, is reduced accordingly, while
the idle planet requires only $0.6$ times the planet face width.}

\subsection{\diff{Stiffness modeling of BiPathCPG Gearbox}}\label{pancake_stiffness}

\diff{
The lumped mesh-stiffness formulation of Section~\ref{sec:stiffness_problem} is
applied to the BiPathCPG, with the strain energy now summed over the $sp1$,
$p2r$, $is$-$p2$, $is$-$ip2$ and $ip2$-$r$ meshes and the mesh forces taken
from~\eqref{eq:solution_gamma}. Writing $c_{\mathrm{idle}}$ for the total compliance of
the idle path, $c_{\mathrm{idle}} = 1/k_{is\text{-}p2} + 1/k_{is\text{-}ip2} +
1/k_{ip2\text{-}r}$, and defining
\begin{equation}
    \beta
    =
    \frac{r_{p1}/r_{p2}}
         {1+\gamma},
\end{equation}
the resulting input-referred torsional stiffness is
\begin{equation}
\begin{aligned}
K_{\mathrm{BPCPG,in}} &= n_p r_s^2 \big[ 1/k_{sp1} \\
  &\quad + \beta^2 \big( 1/k_{p2r} + \gamma^2 c_{\mathrm{idle}} \big) \big]^{-1},
\end{aligned}
\end{equation}
with the output-referred stiffness following as
$K_{\mathrm{BPCPG,out}} = G^2 K_{\mathrm{BPCPG,in}}$.
}

\subsubsection{\diff{Stiffness Comparison of CPG and BiPathCPG}}

\diff{
To obtain a simplified analytical comparison between the two architectures,
all gear meshes are assumed to have the same linear mesh stiffness,
$k_{sp1}=k_{p2r}=k_{is-p2}=k_{is-ip2}=k_{ip2-r}=k$, and
$\alpha=r_{p1}/r_{p2}$ is defined. This equal-stiffness assumption is a
simplification, since the actual mesh stiffness depends on the individual gear
geometry and contact conditions. However, it provides a useful first-order
comparison of the relative stiffness of the CPG and BiPathCPG architectures.
The input-referred stiffness of the CPG then becomes
\begin{equation}
    K_{\mathrm{CPG,in}}
    =
    n_p r_s^2 k
    \left(1+\alpha^2\right)^{-1},
\end{equation}
while, for the BiPathCPG with $\gamma=0.6$,
\begin{equation}
\begin{aligned}
    K_{\mathrm{BPCPG,in}}
    &=
    n_p r_s^2 k
    \left[
        1+
        \frac{\left(1+3\gamma^2\right)\alpha^2}
             {\left(1+\gamma\right)^2}
    \right]^{-1}
    \\[2pt]
    &=
    n_p r_s^2 k
    \left[
        1+0.8125\,\alpha^2
    \right]^{-1}.
\end{aligned}
\end{equation}
The stiffness improvement can therefore be expressed as
\begin{equation}
    \frac{K_{\mathrm{BPCPG,in}}}
         {K_{\mathrm{CPG,in}}}
    =
    \frac{1+\alpha^2}
    {1+0.8125\,\alpha^2}.
\end{equation}
Since $r_{p1}>r_{p2}$ in a CPG, $\alpha>1$, for which the above ratio is
greater than unity. Thus, under this simplified equal-mesh-stiffness
assumption, the BiPathCPG exhibits higher mesh-induced transmission
stiffness than the conventional CPG. For the fabricated design of
Table~\ref{tab:actuator_designs}, where $N_{p1}=66$ and $N_{p2}=33$ share a common module, $\alpha = r_{p1}/r_{p2} = 2$, and the above ratio evaluates to $1.18$, so the BiPathCPG is predicted to be $1.18\times$ stiffer than the conventional CPG.
}

\section{Optimization and CAD Design}
The many interdependent design parameters of CPG and \diff{BiPathCPG} architectures, together with the
additional geometric and meshing constraints of the \diff{BiPathCPG}, make manual parameter
selection difficult and error-prone, motivating a computational design framework. We
therefore extend the COMPAct framework~\cite{singh2025compactcomputationaloptimizationautomated}
to the \diff{BiPathCPG}. Following the same work, the optimizer minimizes the cost function:

\begin{equation}
    C_{\text{act}} = K_m M_{\text{act}} - K_e \eta + K_w W_p
    + K_g \left|GR_{\text{req}} - GR\right|,
\end{equation}
% where $M_{\text{act}}$, $\eta$, $W_p$, and $GR$ denote actuator mass, mechanical
% efficiency, axial width, and gear ratio, respectively, and $K_m$, $K_e$,
% $K_w$, $K_g$ are weighting coefficients.
% \textcolor{black}{The expressions for $\eta$ and $GR$ are given in
% Eqs.~\ref{eq:cpg_eff} and \ref{eq:CPG_ratio}, respectively. The values of
% $M_{\text{act}}$ and $W_p$ are calculated using the detailed mass and geometric model
% presented in \cite{singh2025compactcomputationaloptimizationautomated}.
% The weighting coefficients are set to $K_m = 1~\text{kg}^{-1}$, $K_e = 2$,
% and $K_w = 0.2~\text{mm}^{-1}$, balancing the mass, efficiency, and width
% terms for the actuator sizes considered. The gear-ratio term is active only
% when a target ratio is specified ($K_g = 10$ if $GR_{\text{req}} \neq 0$,
% else $K_g = 0$).}
where $M_{\text{act}}$, $\eta$, $W_p$ and $GR$ are the actuator mass, mechanical
efficiency, axial width and gear ratio, and $K_m$, $K_e$, $K_w$, $K_g$ are
weighting coefficients.
\diff{$\eta$ and $GR$ follow Eqs.~\ref{eq:cpg_eff}
and \ref{eq:CPG_ratio}, and $M_{\text{act}}$ and $W_p$ the mass and geometric
model of~\cite{singh2025compactcomputationaloptimizationautomated}. We set
$K_m = 1~\text{kg}^{-1}$, $K_e = 2$ and $K_w = 0.2~\text{mm}^{-1}$, balancing
the three terms over the actuator sizes considered; $K_g = 10$ when a target
ratio is specified ($GR_{\text{req}} \neq 0$) and $0$ otherwise.}

\begin{figure}[htbp]
    \centering
    \includegraphics[width=1\linewidth]{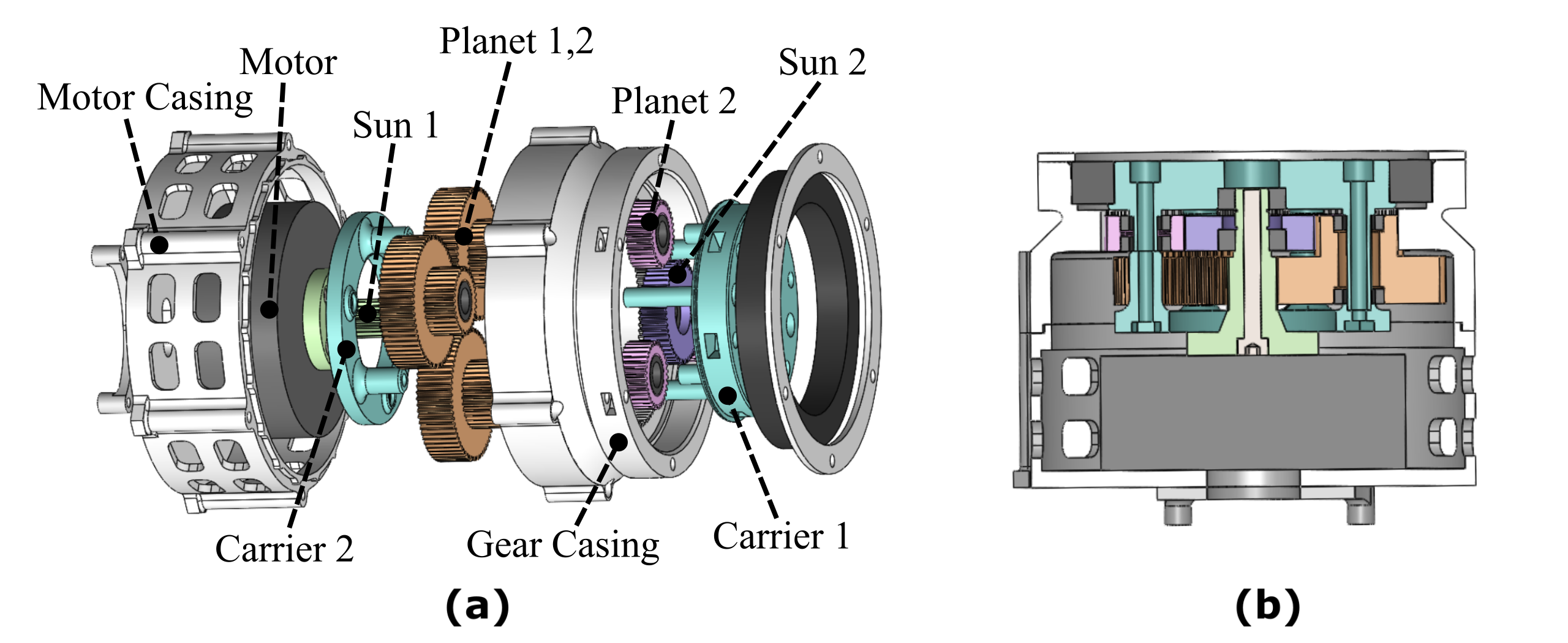}
    \caption{\textbf{\diff{BiPathCPG}}: (a) exploded view, (b) sectional view.}
    \label{fig:pancake_CAD}
\end{figure}

The major changes and results of this framework are explained below.

\begin{figure*}[!t]
    \centering
    \includegraphics[width=0.9\textwidth]{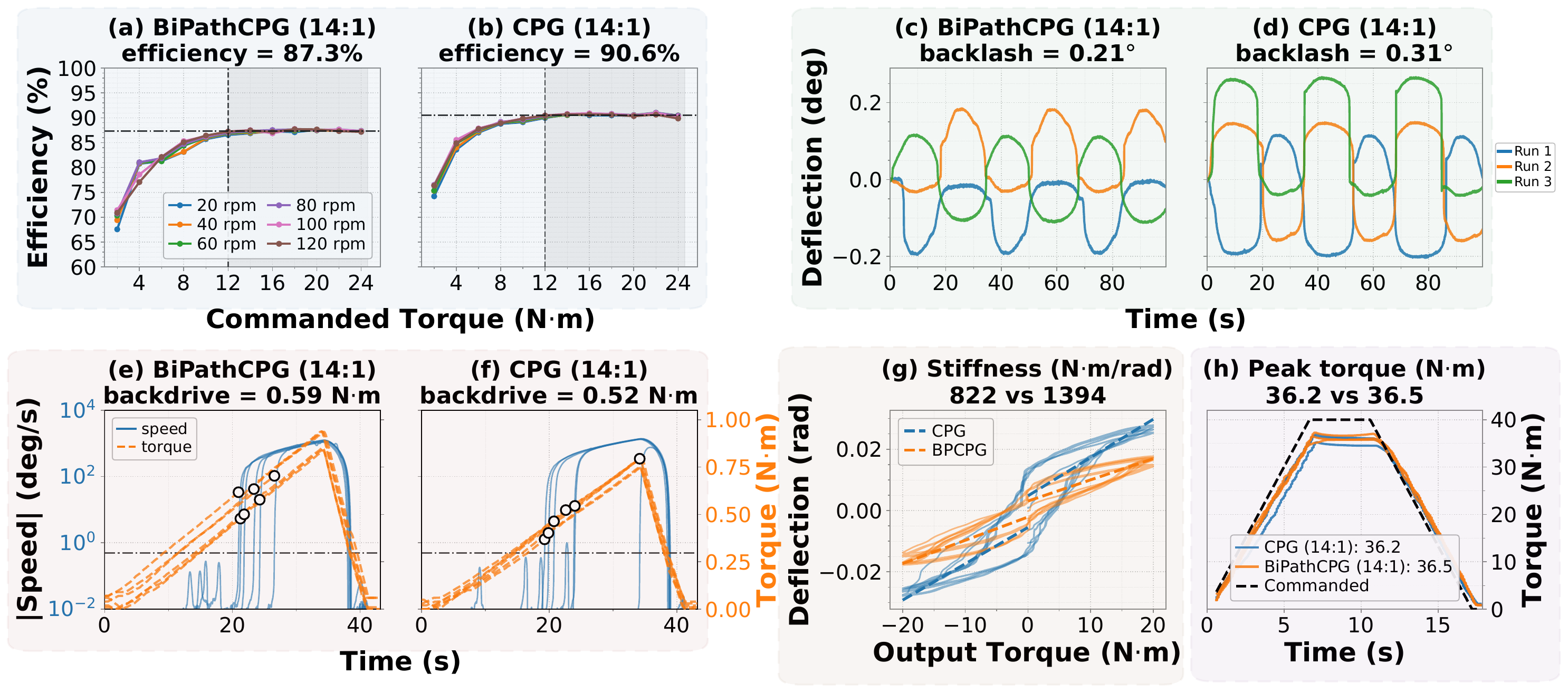}
    \caption{\diff{\textbf{Bench comparison of the CPG and BiPathCPG at $i \approx 14{:}1$.}
    (a), (b) Mechanical efficiency versus commanded torque at $20$--$120$~rpm; the
    shaded band marks the saturated region above $12$~N$\cdot$m and the dash-dot
    line its mean. (c), (d) Output deflection under a $\pm 2$~N$\cdot$m
    fixed-output load cycle. (e), (f) Static load-cell torque during the
    backdriving ramp, with circles marking motion onset. (g) Output deflection
    versus torque with forward and backward linear fits. (h) Load-cell torque
    for a $40$~N$\cdot$m command.}}
    \label{fig:results_grid}
\end{figure*}
\subsubsection{Constraint Modifications}\label{sec:constraints}
In the conventional CPG the second-stage planet count equals that of the first stage,
$n^p_2 = n^p_1$, whereas the \diff{BiPathCPG} doubles it, $n^p_2 = 2n^p_1$. This change,
along with the addition of a stage-2 sun gear, propagates into three classes of
constraints.

\noindent\textbf{a) Geometric Constraints.}\quad
The first geometric relation is identical in both configurations:
\begin{equation}
    m_2 N^r_2 = m_1\!\left(N^s_1 + N^p_1\right) + m_2 N^p_2.
\end{equation}
The \diff{BiPathCPG} additionally enforces the stage-2 geometric constraint:
\begin{equation}
    N^r_2 = N^s_2 + 2N^p_2,
\end{equation}
a condition absent in the conventional CPG due to the lack of a stage-2 sun gear.

\noindent\textbf{b) Meshing Constraints.}\quad
The conventional CPG requires $N^r_2 \bmod n^p_1 = 0$ for proper meshing, while the \diff{BiPathCPG} imposes
the stricter conditions
\begin{equation}\label{meshing}
    N^r_2 \bmod n^p_1 = 0, \qquad
    \left(N^s_2 + N^r_2\right) \bmod 2n^p_1 = 0,
\end{equation}
where the second condition is the standard sun--planet--ring meshing compatibility
applied to stage~2 with $2n^p_1$ equally spaced planets.

\noindent\textbf{c) No Interference Constraints.}\quad
The stage-1 planet-to-carrier clearance constraint is unchanged:
\begin{equation}
    2\!\left(R^s_1 + R^p_1\right)\sin\!\left(\frac{\pi}{2n^p_1}\right)
    - R^p_1 - R_{ce} \ge \delta_{clr}.
\end{equation}
The \diff{BiPathCPG} additionally enforces the corresponding stage-2 clearance constraint:
\begin{equation} \label{no_int}
    2\!\left(R^s_2 + R^p_2\right)\sin\!\left(\frac{\pi}{4n^p_1}\right)
    - R^p_2 \ge \delta_{clr}.
\end{equation}
Here, $R^s_i$ and $R^p_i$ denote the pitch radii of the sun and planet gears in stage~$i$, $R_{ce}=4\text{ mm}$ is the carrier extrusion radius, and $\delta_{clr}=1\text{ mm}$ is the minimum clearance to avoid interference. Note: The no-interference (Eq.~\ref{no_int}) and meshing (Eq.~\ref{meshing})
constraints can lead to radially larger designs.

\subsubsection{Optimization Results and CAD Design}\label{cad+opt}
Two simulation experiments were conducted \diff{for every motor in our
dataset}, across gear ratios from $4{:}1$ to $20{:}1$ in unit increments. The
first used 3D-printed material properties (PLA:
$\rho = 1430~\text{kg/m}^3$, $\sigma_u = 70~\text{MPa}$); the second assigned steel
to gears ($\rho = 7850~\text{kg/m}^3$, $\sigma_u = 400~\text{MPa}$) and aluminum to
remaining components ($\rho = 2710~\text{kg/m}^3$). Both
experiments used identical constraints: $D_{\max} = 1.25\,D_{\text{motor}}$,
$m_{\min} = 0.5~\text{mm}$, and $N_{\min} = 18$.
\diff{The full sweep is shown in the supplementary video; from these
results we selected the \href{https://mad-motor.com/products/mad-components-m6c12-eee-industrial-drone-motor?VariantsId=10508}{MAD M6C12} motor at $i \approx 14{:}1$ for fabrication.}

% \begin{table}[t]
% \caption{Optimized Design Parameters at $i \approx 14{:}1$}
% \label{tab:actuator_designs}
% \centering
% \setlength{\tabcolsep}{3pt}
% \small
% \begin{tabular}{|l|c|c|}
% \hline
% \textbf{Parameter} & \textbf{CPG} & \textbf{P-CPG} \\
% \hline
% $i$                                  & 14.00        & 14.00          \\
% $m$ (mm)                             & 0.6          & 0.6            \\
% $N_s/N_{p1}/N_{p2}/N_r/$\textcolor{black}{$N_{s2}$}              & 18/66/33/117/- & 18/66/33/117/\textcolor{black}{51}   \\
% $n_p$                                & 3            & 3              \\
% $\mathbf{fw}_{[s,p1,p2,r,\textcolor{black}{s2}]}$ (mm)     & [11.2, 11.2, 16.4, 16.4, -] & [11.2, 11.2, \textcolor{black}{\textbf{10.2, 10.2}, 10.2}] \\
% $W_p$ (mm)                             & 87.10        & \textbf{\textcolor{black}{80.19}} \\
% $M_{\text{act}}$ (kg)                & 0.815        & \textbf{\textcolor{black}{0.806}} \\
% $\tau_{\max}$ (N$\cdot$m) \textcolor{black}{(Pred.)}           & \textcolor{black}{36.96}        & \textcolor{black}{36.96}          \\
% efficiency ($\eta$)                         & 0.938        & 0.938          \\
% $\tau/M$ (N$\cdot$m/kg)              & \textcolor{black}{45.35}        & \textcolor{black}{\textbf{45.86}} \\
% \hline
% \end{tabular}
% \end{table}

\begin{table}[t]
\caption{Optimized Parameters, $i\approx14{:}1$\diff{, MAD M6C12}}
\label{tab:actuator_designs}
\centering
\setlength{\tabcolsep}{2.5pt}
\renewcommand{\arraystretch}{1.15}
\footnotesize
\begin{tabular}{|@{\hspace{2pt}}l@{\hspace{2pt}}|c|c|}
\hline
\textbf{Parameter} & \textbf{CPG} & \textbf{\diff{BiPathCPG}} \\
\hline
$i$                                   & 14.00 & 14.00 \\
$m$ (mm)                              & 0.6   & 0.6   \\
$N_s/N_{p1}/N_{p2}/N_r/$\diff{$N_{is2}$}
  & 18/66/33/117/--
  & 18/66/33/117/\diff{51} \\
$n_p$                                 & 3     & 3     \\
\makecell[l]{$fw_{s/p1/p2/r/}$\diff{$_{is2/ip2}$}\\(mm)}
  & \makecell{11.2/11.2/16.4/\\16.4/--/--}
  & \makecell{11.2/11.2/\diff{\textbf{10.2/}}\\\diff{\textbf{10.2/10.2/8.2$^\dagger$}}} \\
$W_p$ (mm)                            & 87.10 & \diff{\textbf{80.94}} \\
$M_{\text{act}}$ (kg)                 & 0.815 & \diff{\textbf{0.801}} \\
$\diff{\tau_{stall}}$ (N$\cdot$m) \diff{(Pred.)}
  & \diff{36.96} & \diff{36.96} \\
$\eta$ \diff{(Theoretical)}                               & 0.938 & 0.938 \\
$\tau/M$ (N$\cdot$m/kg)               & \diff{45.35}
  & \diff{\textbf{46.14}} \\
\hline
\end{tabular}

\vspace{2pt}
{\raggedright\scriptsize
\diff{$^\dagger$The idle-planet face width implied by $\gamma=0.6$ is $0.6\times10.2=6.1$~mm; $8.2$~mm is used, this being the minimum width that accommodates the bearing.}\par}

\end{table}

% A gear ratio of $i \approx 14{:}1$ was designed (Fig. \ref{fig:pancake_CAD}) and selected for fabrication and experimental validation (Table~\ref{tab:actuator_designs}, Fig.~\ref{fig:intro}). The predicted axial widths (87.1, 78.9\,mm) and masses (0.815, 0.789\,kg) for the CPG and Pancake-CPG agree closely with measured values of (85.1, 76.4\,mm) and (0.808, 0.780\,kg), validating the geometry and mass model. The Pancake-CPG is 28\,g lighter with an 8.7\,mm reduction in axial width while maintaining identical gear ratio, efficiency, and peak torque.

\textcolor{black}{The fabricated design is shown in Fig.~\ref{fig:pancake_CAD},
with its parameters in Table~\ref{tab:actuator_designs} and the assembled
actuators in Fig.~\ref{fig:intro}.} The predicted axial
widths \diff{($87.1$, $80.9$\,mm) and masses ($0.815$, $0.801$\,kg)} for the CPG and
\diff{BiPathCPG} agree closely with measured values of \diff{($85.3$, $78.9$\,mm)} and
\diff{($0.820$, $0.795$\,kg)}, validating the geometry and
mass model. The fabricated \diff{BiPathCPG} is \diff{$25$}\,g lighter with a
\diff{$6.4$}\,mm reduction in axial width.

% \begin{figure}[htbp]
%     \centering
%     \includegraphics[width=0.8\linewidth]{images/metal_mass_vs_gear_ratio_all_motors.png}
%     \caption{Gear Ratio vs Actuator Mass comparison for metal actuators}
%     \label{fig:mass_vs_gr}
% \end{figure}

\section{Results}

Based on the optimisation results, a gear ratio of $i = 14{:}1$ (Table~\ref{tab:actuator_designs}) 
with the MAD M6C12 150KV motor was selected for fabrication. Both the CPG and \diff{BiPathCPG} were fully
3D-printed in PLA at 100\% infill using a Bambu Lab X1C printer. All experiments were
conducted using a custom, in-house developed BLDC motor driver.

\subsection{Efficiency Comparison}

Power efficiency was measured using a custom dynamometer that couples the test and load 
actuators output-to-output. A static reaction torque sensor (STARK EMBASYS: SE-TS-F) is 
mounted between the test actuator housing and a rigid ground fixture. The load actuator 
maintains constant angular velocity while the test actuator runs in torque-control mode; 
output power is reaction torque times encoder-derived velocity (scaled by gear ratio), 
and input power is commanded motor torque times motor velocity. Mechanical efficiency is 
the output-input ratio. The setup follows prior 
work~\cite{singh2025compactcomputationaloptimizationautomated, chi2025demonstratingberkeleyhumanoidlite}, 
with a static torque sensor replacing the load cell. 
\diff{The results of this experiment are shown in
Fig.~\ref{fig:results_grid}(a) and (b). The conventional CPG achieves a mechanical
efficiency of approximately $90.6 \pm 0.3\%$ at higher torques, whereas the
BiPathCPG achieves approximately $87.3 \pm 0.3\%$. The lower efficiency of the
BiPathCPG can be attributed to the presence of additional idle planet
gears in the second stage, which increases friction compared with the
conventional CPG. The theoretically modeled efficiency values reported in
Table~\ref{tab:actuator_designs} are the same for both gearboxes because the
efficiency model used in this work~\cite{effModel3K} does not account for the
number of planet gears. Nevertheless, the difference in mechanical efficiency
between the two gearbox architectures remains relatively small.}

\begin{table*}[t]
\centering
\caption{Actuator Performance Comparison}
\label{tab:general_comparison}
\setlength{\tabcolsep}{4pt}
\renewcommand{\arraystretch}{1.1}
\small

\resizebox{\textwidth}{!}{%
\begin{tabular}{@{}l l l c c c c c c c@{}}
\toprule
\textbf{Actuator} &
\diff{\textbf{Motor (Peak Torque)}} &
\diff{\textbf{Gearbox-type (gear material)}} &
\textbf{Ratio} &
\makecell{\textbf{Stiffness}\\(N$\cdot$m/rad)} &
\makecell{\textbf{Backlash}$^\P$\\($^\circ$)} &
\makecell{\textbf{Mass}\\(g)} &
\makecell{\diff{\textbf{D$\times$L (Lg)}}\\\diff{(mm)}} &
\makecell{\diff{\textbf{Exp. Peak Torque}}\\\diff{(N$\cdot$m)}} &
\makecell{\diff{\textbf{Pred. Peak Torque}$^\ast$}\\\diff{(N$\cdot$m)}} \\
\midrule

% ==================== Cycloidal ====================

Roozing Cycloidal~\cite{roozing2022cycloidal}
& \diff{MN8012 (3.82 N$\cdot$m)}
& Cycloidal (PLA)
& 11
& 633
& 0.35
& 809
& \diff{$104\times58 (23)$}
& $>40$
& \diff{42.02}
\\

Roozing (Non-Pinwheel)~\cite{pinwheel}
& \diff{MN8012 (3.82 N$\cdot$m)}
& \diff{Cycloidal (PA6-CF)}
& 11
& 1468
& 0.25
& \diff{797}
& \diff{$104\times58 (23)$}
& 36.4
& \diff{42.02}
\\

\diff{Roozing Anti-backlash~\cite{anti_backlash_cyc}}
& \diff{RO100 (12 N$\cdot$m)}
& \diff{Cycloidal (PA6-CF)}
& \diff{13}
& \diff{300$^\S$}
& \diff{0.647$^\S$}
& \diff{1350}
& \diff{$125\times62 (25)$}
& \diff{156}
& \diff{156}
\\

Berkeley Humanoid Lite~\cite{chi2025demonstratingberkeleyhumanoidlite}
& \diff{M6C12 (2.64 N$\cdot$m)}
& Cycloidal (PLA)
& 15
& 319
& \diff{$1.312 \pm 0.241$}
& ---
& \diff{$88\times66 (35)$}
& \diff{---}
& \diff{39.60}
\\

\diff{CY30~\cite{bezzini2024gearboxes}}
& \diff{Maxon EC90 Flat (5.03 N$\cdot$m)}
& \diff{Cycloidal (PLA)}
& \diff{30}
& \diff{641.4}
& \diff{$1.85 \pm 0.39$}
& \diff{942$^\dagger$}
& \diff{$92\times85.8^\ddagger (33.5)$}
& \diff{---}
& \diff{150.90}
\\

% CY80v1~\cite{bezzini2024gearboxes}
% & \textcolor{black}{Maxon EC 90 Flat, 360 W (5.03 N$\cdot$m)}
% & Cycloidal (PLA)
% & 80
% & 278.1
% & 8.4099
% & \textcolor{black}{822$^\dagger$}
% & \textcolor{black}{---}
% & \textcolor{black}{402.40}
% \\

% CY80v2~\cite{bezzini2024gearboxes}
% & \textcolor{black}{Maxon EC 90 Flat, 360 W (5.03 N$\cdot$m)}
% & Cycloidal (PLA)
% & 80
% & 659.5
% & 2.69
% & \textcolor{black}{946$^\dagger$}
% & \textcolor{black}{---}
% & \textcolor{black}{402.40}
% \\

% CY80v3~\cite{bezzini2024gearboxes}
% & \textcolor{black}{Maxon EC 90 Flat, 360 W (5.03 N$\cdot$m)}
% & Cycloidal (PLA)
% & 80
% & 344.7
% & 1.89
% & \textcolor{black}{895$^\dagger$}
% & \textcolor{black}{---}
% & \textcolor{black}{402.40}
% \\

\midrule

% ==================== Planetary / WPG / CPG ====================

COMPAct SSPG~\cite{singh2025compactcomputationaloptimizationautomated}
& \diff{MN8014 (4.97 N$\cdot$m)}
& \diff{Planetary (single stage) (PLA)}
& 7.2
& 242.7
& 0.6
& 838
& \diff{$102\times82 (47)$}
& \diff{---}
& \diff{35.78}
\\

EMBIR SSPG~\cite{embiract}
& \diff{RI50 (1.67 N$\cdot$m)}
& \diff{Planetary (single stage) (PLA)}
& 7.5
& ---
& ---
& 482
& \diff{$90\times70 (\text{---})$}
& \diff{7.5(20s)/19.1(2s)}
& \diff{12.53}
\\

EMBIR WPG~\cite{embiract}
& \diff{RI50 (1.67 N$\cdot$m)}
& \diff{Planetary (Wolfrom/3K) (PLA)}
& 15.04
& ---
& ---
& 536
& \diff{$90\times90 (\text{---})$}
& \diff{15.0(20s)/38.2(2s)}
& \diff{25.12}
\\

\diff{CO30~\cite{bezzini2024gearboxes}}
& \diff{Maxon EC90 Flat (5.03 N$\cdot$m)}
& \diff{Planetary (Wolfrom/3K) (PLA)}
& \diff{30}
& \diff{723.2}
& \diff{$0.53 \pm 0.21$}
& \diff{1002$^\dagger$}
& \diff{$100\times105.9^\ddagger (53.6)$}
& \diff{---}
& \diff{150.90}
\\

% CO80~\cite{bezzini2024gearboxes}
% & \textcolor{black}{Maxon EC 90 Flat, 360 W (5.03 N$\cdot$m)}
% & WPG(3K) (PLA)
% & 80
% & 821.7
% & 0.23
% & \textcolor{black}{996$^\dagger$}
% & \textcolor{black}{---}
% & \textcolor{black}{402.40}
% \\

\diff{CFB~\cite{bezzini2026embedded}}
& \diff{Maxon EC90 Flat (5.03 N$\cdot$m)}
& \diff{Planetary (Wolfrom/3K) (PLA)}
& \diff{30}
& \diff{971.5}
& \diff{$0.114 \pm 0.076$}
& \diff{899$^\dagger$}
& \diff{$90\times100.7^\ddagger (48.4)$}
& \diff{---}
& \diff{150.90}
\\

% C3D~\cite{bezzini2026embedded}
% & \textcolor{black}{Maxon EC 90 Flat, 360 W (5.03 N$\cdot$m)}
% & Compound Planetary (PLA)
% & 30
% & 653.7
% & 0.195
% & \textcolor{black}{816$^\dagger$}
% & \textcolor{black}{---}
% & \textcolor{black}{150.90}
% \\

\midrule

% ==================== Planetary-Cycloidal Hybrids ====================

% PLACYd~\cite{bezzini2026nested}
% & \textcolor{black}{Maxon EC 90 Flat, 360 W (5.03 N$\cdot$m)}
% & Planetary-Cycloidal
% & 45
% & 859.6
% & 0.121
% & \textcolor{black}{819$^\dagger$}
% & \textcolor{black}{---}
% & \textcolor{black}{226.35}
% \\

\diff{CP30~\cite{bezzini2024gearboxes}}
& \diff{Maxon EC90 Flat (5.03 N$\cdot$m)}
& \diff{Cycloidal-Planetary (PLA)}
& \diff{30}
& \diff{573.6}
& \diff{1.10}
& \diff{910$^\dagger$}
& \diff{$78\times104.3^\ddagger (52.0)$}
& \diff{---}
& \diff{150.90}
\\

\diff{PLACYc~\cite{bezzini2026nested}}
& \diff{Maxon EC90 Flat (5.03 N$\cdot$m)}
& \diff{Planetary-Cycloidal (PLA+N12CF)}
& \diff{45}
& \diff{918.1}
& \diff{$0.066 \pm 0.029$}
& \diff{792$^\dagger$}
& \diff{$70.5\times82.3^\ddagger (30.0)$}
& \diff{---}
& \diff{226.35}
\\

\diff{CYPLO~\cite{bezzini2025cyplo}}
& \diff{Maxon EC90 Flat (5.03 N$\cdot$m)}
& \diff{Cycloidal-Planetary (PLA+N12CF)}
& \diff{52.5}
& \diff{859.6}
& \diff{0.175}
& \diff{909$^\dagger$}
& \diff{$87\times82.75^\ddagger (30.45)$}
& \diff{---}
& \diff{264.08}
\\

% CP80~\cite{bezzini2024gearboxes}
% & \textcolor{black}{Maxon EC 90 Flat, 360 W (5.03 N$\cdot$m)}
% & Cycloidal-Planetary (PLA)
% & 80
% & 472.5
% & 0.8898
% & \textcolor{black}{940$^\dagger$}
% & \textcolor{black}{---}
% & \textcolor{black}{402.40}
% \\

\midrule

CPG (Ours\diff{: Baseline})
& \diff{M6C12 (2.64 N$\cdot$m)}
& CPG (PLA)
& 14
& \diff{822}
& \diff{$0.311 \pm 0.004$}
& \diff{820}
& \diff{$104.7\times85.3 (48.7)$}
& \diff{36.22}
& \diff{36.96}
\\

\diff{BiPathCPG} (Ours\diff{: Proposed})
& \diff{M6C12 (2.64 N$\cdot$m)}
& \diff{BiPathCPG} (PLA)
& 14
& \diff{\textbf{1394}}
& \diff{\textbf{$0.213 \pm 0.014$}}
& \diff{\textbf{795}}
& \diff{\textbf{$104.7\times78.9 (42.3)$}}
& \diff{\textbf{36.49}}
& \diff{\textbf{36.96}}
\\

\bottomrule
\end{tabular}%
}

\vspace{2pt}
{\raggedright\scriptsize
\diff{D$\times$L is the actuator diameter $\times$ axial length; the value in brackets, L$_g$, is the axial length of the gearbox alone. In the CPG and BiPathCPG of this work the motor--gearbox coupling sits outside the gearbox and is therefore counted in L but not in L$_g$; in designs with an internal coupling the coupling is already contained in L$_g$. $^\ast$Predicted peak torque is the manufacturer-specified motor peak torque multiplied by the gearbox ratio. $^\dagger$The referenced Bezzini papers report gearbox mass only; a motor mass of 638~g has been added to obtain the actuator mass. $^\ddagger$The referenced Bezzini papers report gearbox dimensions only; a motor axial width of 52.3~mm has been added to the gearbox height to obtain the actuator axial length. For CYPLO, PLACYc and CP30 the reported gearbox diameter is smaller than the $\phi$90~mm motor, so the actuator diameter is set by the motor rather than the gearbox. $^\S$Stiffness and backlash reported for the conic-disk anti-backlash variant at $100\%$ preload. $^\P$Backlash reporting is not uniform across sources: mean $\pm$ standard deviation where available, and worst-case or typical play otherwise.}\par}

\end{table*}

\subsection{\diff{Backlash and Transmission Stiffness}}
\label{sec:backlash}

\diff{Both quantities were measured on the same fixed-output rig
following \cite{anti_backlash_cyc}, with the output rigidly constrained by a
steel plate fixed to the ground. The applied torque was ramped to the target
over $7$~s, held for $3$~s, and returned to zero, then repeated in the negative
direction, giving a $33$~s load cycle. Each cycle was repeated three times at
each of three output orientations spaced $120^\circ$ apart, giving nine
measurements per gearbox. Backlash was measured at $\pm 2$~N$\cdot$m (about
$0.14$~N$\cdot$m at the motor), low enough that gearbox compliance contributes
negligibly and the deflection is backlash alone; stiffness was measured at
$\pm 20$~N$\cdot$m. Deflection was read from the $16$-bit motor-side encoder,
resolving $0.0055^\circ$ at the motor side.}

\diff{For each measurement, backlash was computed as the difference
between the maximum and minimum output-referred angular deflection over the load
cycle, and the reported value is the average of the nine measurements
(Fig.~\ref{fig:results_grid}(c) and (d)). The measured backlash is
$0.213 \pm 0.014^\circ$ for the BiPathCPG and $0.311 \pm 0.004^\circ$ for the CPG, a
$32\%$ reduction. This primarily results from the increased number of
simultaneously engaged gear meshes in the BiPathCPG.}

Stiffness was estimated by fitting $\theta = \frac{\tau}{K} + b$
to the encoder deflection separately for each direction of loading, taking $K$ as
the mean of the forward and backward fits. \diff{The conventional CPG yielded
$822 \pm 10$~N$\cdot$m/rad ($801$ forward, $843$ backward), while the
BiPathCPG reached $1394 \pm 167$~N$\cdot$m/rad ($1460$ forward, $1328$
backward), a $1.70\times$ increase (Fig.~\ref{fig:results_grid}(g)), above the
$1.18\times$ predicted in Section~\ref{pancake_stiffness}. The difference can be attributed to the fact that the model assumes equal tangential mesh springs, no shaft or gear-body compliance, and 
omits an effect that acts only on the BiPathCPG: under ring load the
stage-2 planets deflect radially inward in the CPG, whereas the idle sun braces
them from within. This likely accounts for much of the difference.}
The \diff{BiPathCPG}'s stiffness exceeds the Berkeley Humanoid Lite and COMPAct
actuators by more than \diff{$4\times$, and reaches $95\%$ of the
PA6-CF non-pinwheel cycloidal actuator~\cite{pinwheel}
(Table~\ref{tab:general_comparison})}, despite relying on spur-gear meshing
rather than the inherently stiffer rolling contact of cycloidal drives.

\subsection{Stall Peak Torque Test}
We evaluate stall peak torque with a custom rig: the actuator is rigidly mounted to a fixed base, 
and a $30$\,cm radial arm on the output shaft presses against a load cell at its distal end. 
Output torque is the product of the measured contact force and the arm length.During the 
experiment, the commanded torque is increased linearly from zero to a predefined maximum value, 
held at the peak for $t = 6\,\mathrm{s}$, and then decreased linearly back to zero. This procedure 
allows us to estimate the stall peak torque of the actuator. Fig.~\ref{fig:results_grid}(h) shows 
the measured torque profiles. 
\diff{The conventional CPG and the BiPathCPG reach peak torques of $36.22 \pm 0.86$ 
and $36.49 \pm 0.74$~N$\cdot$m respectively, averaged over three trials. The two architectures 
are therefore equivalent in peak torque capacity, with the BiPathCPG matching the conventional 
CPG despite its lower mass and axial width.}

\subsection{\diff{Backdrivability Test}}
\diff{Backdrivability was measured on the same setup used
for the efficiency tests, with a load motor mounted to a static torque
cell and coupled to the output of the actuator under test. Before each trial, the actuator
was rotated in the loading direction so that the gear teeth were already in
contact and any backlash was taken up. The load motor then ramped the output
torque from $0$ to $1.0$~N$\cdot$m over $25$~s, while load position and
torque-cell readings were logged. The test was repeated three times clockwise
and three times counter-clockwise for each gearbox. The
actuator was considered back-driven at the first instant where the load
speed, averaged over $0.5$~s, exceeded $0.5^\circ$/s and remained above this
threshold for $10$ consecutive samples; the back-drive torque is the
load-cell torque averaged over the $0.5$~s window at that instant. The
results are shown in Fig.~\ref{fig:results_grid}(e) and (f). The mean
back-drive torque is $0.59$~N$\cdot$m for the BiPathCPG ($0.65$~N$\cdot$m
CW, $0.52$~N$\cdot$m CCW) and $0.52$~N$\cdot$m for the CPG ($0.60$~N$\cdot$m
CW, $0.43$~N$\cdot$m CCW). As expected from the efficiency tests, the
BiPathCPG is slightly harder to back-drive than the CPG.}

\subsection{\diff{Steady-State Operation and Failure Tests}}
\label{sec:endurance}

\diff{Durability was evaluated using steady-state and failure 
cyclic tests. In both, the actuator oscillated between $+45^\circ$ and $-90^\circ$ 
while driving a $1$~kg payload. In the steady-state test, the payload was mounted on 
a $30$~cm extrusion and oscillated at $0.3$~Hz, producing a measured peak torque of 
approximately $8$~N$\cdot$m per cycle. Both gearboxes operated for $12$~hours without 
failure and showed negligible wear.
In the failure test, the payload was mounted on a $75$~cm extrusion and oscillated at 
$0.34$~Hz. The quasi-static gravitational torque was approximately $11$~N$\cdot$m, while 
arm inertia produced additional spikes at direction reversals, raising the measured peak 
torque to approximately $28$~N$\cdot$m. Results of both tests and representative 
torque-versus-time profiles are provided in the supplementary video.
Under this loading, the BiPathCPG operated for $45$~minutes before the stage-1 sun 
gear stripped, with negligible wear in the remaining gears. The conventional CPG failed 
similarly after $37$~minutes. This is likely because both designs share an identical 
first stage (Table~\ref{tab:actuator_designs}), with the smallest and most heavily 
loaded stage-1 sun determining the torque-limiting failure mode.}

\subsection{\diff{Actuator Comparison}}

\diff{Table~\ref{tab:general_comparison} positions the BiPathCPG
among axially compact 3D-printed gearboxes. Its transmission stiffness is the
second highest listed and the highest of any all-PLA design. Its backlash is
lower than that of every pure cycloidal entry; only designs operating at
substantially higher gear ratios achieve lower values. In axial width the
cycloidal designs are the clear winners at comparable gear ratios, and they
remain competitive at much higher ones, reaching reductions of $30{:}1$ and
above within a similar axial envelope. For mass the fairer comparison is torque
density, which is similar to that of pure cycloidal gearboxes at comparable gear
ratios; because cycloidal designs reach high reductions in very compact volumes,
however, matching their torque density at high gear ratios is not attainable
with the BiPathCPG. Against its own CPG baseline the BiPathCPG nevertheless
improves stiffness, backlash, mass and axial width, as reported above.}

\section{Conclusion}

This work presents the \diff{BiPathCPG}, a novel
architecture that improves conventional compound planetary actuators, particularly
3D-printed ones, by adding \diff{idle} load-sharing planets and a sun gear within the
otherwise unused volume of a CPG. This yields simultaneous gains in torsional
stiffness, width, and mass, validated by fabricating and testing optimizer-selected
\diff{BiPathCPG} and CPG designs on custom-built rigs. Limitations include \diff{lower
efficiency and slightly harder backdriving from the added idle meshes}, and increased assembly complexity due to the additional idle gears. \diff{Under dynamic loading peaking at
$28$~N$\cdot$m, both gearboxes stripped the stage-1 sun within an hour of
cycling.} Despite these limitations, the \diff{BiPathCPG} proves a compelling alternative
to CPG actuators for legged robots. Future work includes more durable designs
with \diff{SLA resin suns~\cite{embiract}}, metal implementations, rigorous endurance
testing, and analogous load-sharing in other two-stage
topologies such as the Wolfrom gearbox.

\bibliographystyle{IEEEtran}
\bibliography{references}

\end{document}